\documentclass[final,twocolumn]{elsarticle}

\usepackage{amssymb}
\usepackage{amsthm}

\usepackage{algorithm}   
\usepackage{algpseudocode} 
\algnewcommand\Yield{\textbf{yield }}

\usepackage{lineno}

\usepackage{caption}
\usepackage{subcaption}
\usepackage{booktabs}
\usepackage{nicematrix}
\usepackage{amsmath, amsfonts}
\usepackage{diagbox}
\usepackage{dsfont}
\usepackage{hyperref}
\usepackage{mathabx}

\newcommand{\norm}[1]{\lVert #1 \rVert_2}
\newcommand{\normsq}[1]{\norm{#1}^2}
\newcommand{\mb}[1]{\mathbf{#1}}

\def\bTR{\mb{T}}

\def\X{\mb{X}}
\def\XT{\X^\bTR}
\def\Y{\mb{Y}}
\def\YT{\Y^\bTR}
\def\R{\mb{R}}
\def\Q{\mb{Q}}
\def\W{\mb{W}}
\def\P{\mb{P}}
\def\PT{\P^\bTR}
\def\vr{\mb{r}}        
\def\q{\mb{q}}
\def\qT{\q^\bTR}
\def\w{\mb{w}}
\def\vp{\mb{p}}        
\def\rT{\vr^\bTR}      
\def\pT{\vp^\bTR}      
\def\t{\mb{t}}
\def\tT{\t^\bTR}
\def\T{\mb{T}}
\def\I{\mb{I}}

\def\zeros{\mb{0}}
\def\zerosT{\zeros^\bTR}

\newcommand{\mysecref}[1]{Section~\ref{#1}}
\newcommand{\myfigref}[1]{Figure~\ref{#1}}
\newcommand{\myeqnref}[1]{Equation~(\ref{#1})}

\newcommand{\mypropref}[1]{Proposition~\ref{#1}}
\newcommand{\mylemmaref}[1]{Lemma~\ref{#1}}

\newcommand{\myalgref}[1]{Algorithm~\ref{#1}}

\newcommand{\mytheoremref}[1]{Theorem~\ref{#1}}

\newtheorem{proposition}{Proposition}
\newtheorem{lemma}{Lemma}
\newtheorem{theorem}{Theorem}
\newtheorem{corollary}{Corollary}

\makeatletter
\def\ps@pprintTitle{%
  \let\@oddhead\@empty
  \let\@evenhead\@empty
  \def\@oddfoot{}%
  \let\@evenfoot\@oddfoot}
\makeatother

\begin{document}

\begin{frontmatter}



\title{Improving Improved Kernel PLS}


\author[1]{Ole-Christian Galbo Engstr{\o}m}

\affiliation[1]{organization={FOSS Analytical A/S},
            addressline={Nils Foss Allé 1}, 
            city={Hiller{\o}d},
            postcode={3400}, 
            country={Denmark}}

\begin{abstract}
Improved Kernel Partial Least Squares (IKPLS) algorithms 1 and 2 are among the fastest PLS calibration algorithms. This article focuses on two shared steps, the computation of the $\X$ rotations, $\R$, and the $\Y$ loadings, $\Q$, and accelerates both. For $\R$, term-by-term accumulation is replaced by a direct evaluation strategy that requires the same number of multiplications but parallelizes better on modern hardware. For $\Q$, I identify --- to the best of my knowledge, for the first time --- equivalences showing that each $\Y$ loading is obtainable, up to explicitly derived constants, from quantities already computed earlier in the same iteration, and I exploit them in IKPLS to reduce the cost of each loading from $\Theta\left(KM\right)$ to $\Theta\left(M\right)$ operations whenever $M = 1$ or $2 \leq M < K$, with $K$ predictor variables (number of columns in $\X$) and $M$ response variables (number of columns in $\Y$). Both improvements provably yield exactly the same $\W$, $\P$, $\Q$, $\R$, and $\T$ as the original algorithms. Benchmarks with NumPy (CPU) and JAX (GPU) show speedups of up to two orders of magnitude for the isolated steps and of approximately $2\times$ (CPU) and $6\times$ (GPU) for entire fits. Both improvements are implemented in the free, open-source Python package \texttt{ikpls}.
\end{abstract}

\begin{keyword}
partial least squares \sep computational complexity \sep algorithm design \sep chemometrics
\end{keyword}

\end{frontmatter}


\pagebreak

\section{Introduction}\label{sec:intro}
Partial Least Squares (PLS, \citealt{wold1966estimation}) is a primary tool in chemometrics (\citealt{brereton2018chemometrics, sorensen2021nir}). It is commonly used to analyze near-infrared (NIR) spectra in both regression (\citealt{wold1966estimation, wold2001pls}) and classification (\citealt{sjostrom1986pls, staahle1987partial, barker2003partial, brereton2014partial}).

Preprocessing is commonly applied to NIR spectra prior to PLS calibration. However, the best preprocessing is hard to determine ahead of modeling and thus typically requires experimentation to find (\citealt{rinnan2009review}). Here, chemometricians should be concerned with the speed and stability of their chosen PLS algorithm. While the original NIPALS algorithm (\citealt{wold1966estimation}) is numerically stable (\citealt{andersson2009comparison, bjorck2017fast}), it is also slow (\citealt{alin2009comparison}). Improved Kernel PLS (IKPLS; \citealt{dayal1997improved}) algorithms 1 and 2 are both fast PLS algorithms (\citealt{alin2009comparison, andersson2009comparison}). Algorithm 1 is also numerically stable (\citealt{andersson2009comparison, bjorck2017fast}), except on pathological data (\citealt{bjorck2017fast}), and algorithm 2 has been shown to produce equivalent results in practice (\citealt{engstrom2024fast}). Faster computation was indeed a motivation for the development of IKPLS (\citealt{dayal1997improved}). Thus, IKPLS is a strong choice of PLS algorithm for the chemometrician who requires both calibration precision and speed.

In this article, I identify two steps in both IKPLS algorithms that can be further sped up. They concern the computations of $\\R$ ($\X$ rotations) and $\Q$ ($\Y$ loadings), respectively. I prove the mathematical equivalence between the original IKPLS steps and the improvements introduced here, and I show that the computation of $\Q$ can be sped up by a factor of $\Theta\left(K\right)$ in most cases. Although the improved computation of $\R$ has the same asymptotic runtime as the original computation, both improvements demonstrate improved runtime on datasets of varying sizes and shapes, illustrating their effect in the broader context of executing the full PLS algorithms. I have implemented the improved IKPLS algorithms with NumPy (\citealt{harris2020array}) and JAX (\citealt{jax2018github}) and released them in the free, open-source Python package \texttt{ikpls} (\citealt{engstrom2024fast}), which also features sample-weighted PLS (\citealt{becker2016accounting, engstrom2025near}) and the combination of IKPLS with the fast cross-validation algorithms by \citet{engstrom2025fast}.

The rest of this article is organized as follows. \mysecref{sec:nomenclature} introduces the notation used and is consistent with the notation in the original IKPLS article (\citealt{dayal1997improved}). Sections \ref{sec:improvement_r} and \ref{sec:improvement_q} introduce the two improvements to the IKPLS algorithms and prove both their correctness and, for the improvement in \mysecref{sec:improvement_q}, superior asymptotic runtime. \mysecref{sec:improvements_additional} briefly mentions additional efficient implementation details of IKPLS in \texttt{ikpls} (\citealt{engstrom2024fast}). \mysecref{sec:benchmarks} exemplifies the faster practical runtime when applying the improvements introduced in Sections \ref{sec:improvement_r} and \ref{sec:improvement_q}. \mysecref{sec:conclusion} provides concluding remarks.

\section{Nomenclature}\label{sec:nomenclature}
The notation in this work is identical to the one used by \citet{dayal1997improved}, so that readers familiar with that work may easily compare it to this work. For convenience, the nomenclature from \citet{dayal1997improved} is written below.

\begin{tabular}{@{}p{1.5cm}p{\dimexpr\linewidth-1.5cm-2\tabcolsep\relax}@{}}
    $\X$        & predictor variables matrix ($N \times K$). \\
    $\Y$        & response variables matrix ($N \times M$). \\
    $\W$        & PLS weights matrix for $\X$ ($K \times A$). \\
    $\P$        & PLS loadings matrix for $\X$ ($K \times A$). \\
    $\Q$        & PLS loadings matrix for $\Y$ ($M \times A$). \\
    $\R$        & PLS weights matrix to compute scores $\mb{T}$ directly from original $\X$ ($K \times A$). \\
    $\T$        & PLS scores matrix of $\X$ ($N \times A$). \\
    $\w_a$      & a column vector of $\mb{W}$. \\
    $\vp_a$     & a column vector of $\mb{P}$. \\
    $\q_a$      & a column vector of $\Q$. \\
    $\vr_a$     & a column vector of $\R$. \\
    $\t_a$      & a column vector of $\T$. \\
    $K$         & number of $\X$-variables. \\
    $M$         & number of $\Y$-variables. \\
    $N$         & number of objects. \\
    $A$         & number of components in PLS model. \\
    $a$         & integer counter for latent variable dimension. \\
\end{tabular}

This work will denote a matrix with subscript $a$ as the state of the matrix at latent-variable dimension $a$. E.g., $\P_a$ is the PLS loadings matrix for $\X$ with the first $a$ latent dimensions, and thus it has $K$ rows and $a$ columns. In contrast, $\left(\XT\Y\right)_a$ denotes the matrix product $\XT\Y$ deflated with the first $a-1$ components, and thus it has $K$ rows and $M$ columns. Similarly, $\left(\YT\X\XT\Y\right)_a$ and $\left(\XT\Y\YT\X\right)_a$ denote $\left(\XT\Y\right)_a^\bTR\left(\XT\Y\right)_a$ and $\left(\XT\Y\right)_a\left(\XT\Y\right)_a^\bTR$, respectively.

In addition to this notation, $\zeros_{*}$ denotes a length $*$ column vector of zeros, and $\delta$ is the Kronecker delta.

Finally, PLS1 refers to the case when $M=1$ and PLS2 refers to the case when $M \geq 2$, as is standard in the chemometric literature.

\section{Improved computation of \texorpdfstring{$\R$}{R
}}\label{sec:improvement_r}
The original step 3 for computation of $\R$ in both IKPLS algorithms (\citealt{dayal1997improved}) is due to \citet{hoskuldsson1988pls} and is given by the following recurrence relation.
\begin{equation}\label{eq:r_original}
    \vr_a = \begin{cases}
        \w_a & a = 1 \\
        \w_a - \sum_{i=1}^{a-1} \left(\pT_i\w_a\right)\vr_i & a \geq 2
    \end{cases}
\end{equation}
As written, \myeqnref{eq:r_original} invites evaluating the sum by term-by-term accumulation over the $a-1$ preceding components, and this is how the implementation in the appendix of \citet{dayal1997improved} proceeds. The accumulation is not necessary, however: \citet[Equations~19--20]{dejong1993}, who attributes it to \citet{hoskuldsson1992}, gives a direct computation of $\vr_a$ as a single matrix-vector product with an explicit $K \times K$ matrix maintained across components by a rank-one update. Their formulation requires $\Theta\left(K^2\right)$ multiplications per component and an additional $\Theta\left(K^2\right)$ memory, whereas \myeqnref{eq:r_original} requires just $\Theta\left(K(a-1)\right)$ multiplications per component (\mypropref{prop:runtime_r_orig}). Since $a \leq K$ in PLS, and typically $a \ll K$, this difference of a factor $\Theta\left(K/a\right)$ is substantial. Consider now an alternative, direct approach to computing $\vr_a$ that requires no such matrix but only the $\R_{a-1}$ and $\P_{a-1}$ that IKPLS accumulates in any case. It is given by

\begin{equation}\label{eq:r_improved}
    \vr_a = \begin{cases}
        \w_a & a = 1 \\
        \w_a - \R_{a-1} \left(\PT_{a-1} \w_a\right) & a \geq 2
    \end{cases}
\end{equation}

The evaluation strategy of \myeqnref{eq:r_improved} appears, without statement or proof, in the function \texttt{kernelpls.fit} of the R package \texttt{pls} \citep[version~2.9-0]{liland2026pls, mevik2007the}, an implementation of IKPLS algorithm~1, which uses it for $a > 5$ and falls back to term-by-term accumulation for $2 \leq a \leq 5$.

\subsection{Correctness}

I will now prove the correctness of \myeqnref{eq:r_improved}.

\begin{proposition}[Correctness, \myeqnref{eq:r_improved}]\label{prop:r_correctness}
    \myeqnref{eq:r_original} and \myeqnref{eq:r_improved} compute the same value of $\vr_a$.

    \begin{proof}
        Recall from \citet{dayal1997improved} that $\R_a = \begin{bmatrix} \vr_1 & \vr_2 & \cdots & \vr_a \end{bmatrix}$ and $\P_a = \begin{bmatrix} \vp_1 & \vp_2 & \cdots & \vp_a \end{bmatrix}$. Then, in the case of $a=1$, the equations are identical. In the case of $a \geq 2$, write out the matrix vector product to obtain
        
        \begin{equation}\label{eq:r_equivalence}
            \begin{split}
                \R_{a-1} \left(\PT_{a-1} \w_a\right) &= \R_{a-1} \begin{bmatrix} \pT_1 \w_a \\ \pT_2 \w_a \\ \vdots \\ \pT_{a-1}\w_a \end{bmatrix}\\
                &= \begin{bmatrix} \vr_1 & \vr_2 & \cdots & \vr_{a-1} \end{bmatrix} \begin{bmatrix} \pT_1 \w_a \\ \pT_2 \w_a \\ \vdots \\ \pT_{a-1}\w_a \end{bmatrix}\\
                &= \sum_{i=1}^{a-1} \vr_i \left(\pT_i\w_a\right)\\
                &= \sum_{i=1}^{a-1} \left(\pT_i\w_a\right)\vr_i
            \end{split}
        \end{equation}
        
        Thus, in the case of $a \geq 2$, both \myeqnref{eq:r_original} and \myeqnref{eq:r_improved} subtract the term in \myeqnref{eq:r_equivalence} from $\w_a$ to obtain $\vr_a$, finalizing the proof that the two equations are equivalent.
    \end{proof}
\end{proposition}

\subsection{Runtime}
To analyze the computational complexity of computing $\vr$ via \myeqnref{eq:r_original} and \myeqnref{eq:r_improved}, I ignore the summations and subtractions and focus solely on the number of multiplication operations required, as multiplication dominates both in number of operations and the practical constant of execution on modern hardware.

\begin{proposition}\label{prop:runtime_r_orig}
    \myeqnref{eq:r_original} requires  $2K(a-1)$ multiplications to compute $\vr_a$.
    \begin{proof}
        When $a=1$, \myeqnref{eq:r_original} returns $\w_a$ directly, requiring $2K(a-1)=0$ multiplications. When $a \geq 2$, computing the inner vector product in the parentheses requires $K$ multiplications, and so does the subsequent scalar vector multiplication, for a total of $2K$ multiplications. Summing over $a-1$ products means the total number of multiplications reaches $2K(a-1)$.
    \end{proof}
\end{proposition}

\begin{proposition}\label{prop:runtime_r_improved}
    \myeqnref{eq:r_improved} requires  $2K(a-1)$ multiplications to compute $\vr_a$.
    \begin{proof}
        When $a=1$, \myeqnref{eq:r_improved} returns $\w_a$ directly, requiring $2K(a-1)=0$ multiplications. When $a \geq 2$, the matrix-vector product in the parentheses must be evaluated first, as this yields the fewest total multiplications. $\PT_{a-1}\w_a$ is a multiplication between a matrix of size $(a-1) \times K$ and a vector of size $K \times 1$, requiring $K \times (a-1)$ multiplications to yield a vector of size $(a-1) \times 1$, which is then left-multiplied by $\R_{a-1}$ of size $K \times (a-1)$ requiring an additional $K \times (a-1)$ multiplications for a total of $2K(a-1)$ multiplications.
    \end{proof}
\end{proposition}

\begin{proposition}\label{prop:runtime_r_equal}
    \myeqnref{eq:r_original} and \myeqnref{eq:r_improved} require the same number of multiplications to compute $\vr_a$.
    \begin{proof}
        This follows directly from \mypropref{prop:runtime_r_orig} and \mypropref{prop:runtime_r_improved}.
    \end{proof}
\end{proposition}

Therefore, the gain from swapping the sequential \myeqnref{eq:r_original} for the direct \myeqnref{eq:r_improved} is solely due to the matrix products being better suited for parallel execution on modern multi-processor hardware.

\section{Improved computation of \texorpdfstring{$\Q$}{Q}}\label{sec:improvement_q}
The improvement for computation of $\vr_a$ shown in \mysecref{sec:improvement_r} is always applicable when $A \geq 2$. In contrast, the improvement in the computation of $\q_a$ shown in this section applies only when $M=1$ (commonly referred to as PLS1) or $2 \leq M < K$ (some cases of PLS2). Thus, the improved computation of $\q_a$ does not apply when $2 \leq M \land K \leq M$, meaning $\Y$ has at least two columns and at least as many columns as $\X$ (the remaining cases of PLS2).

First, consider step 2 of the IKPLS algorithms as given by \citet{dayal1997improved}, shown in \myalgref{alg:step_2}.

\begin{algorithm}
    \caption{Step 2 of IKPLS}\label{alg:step_2}
    \begin{algorithmic}[1]
    \If{$M = 1$} \Comment{PLS1}
        \State $\tilde{\w}_a \gets \left(\XT\Y\right)_a$
    \Else \Comment{PLS2}
        \If{$M < K$} \Comment{$2 \leq M < K$}
            \State $\lambda_a \gets \text{largest eigenvalue of } \left(\YT\X\XT\Y\right)_a$
            \State $\tilde{\q}_a \gets \text{eigenvector corresponding to } \lambda_a$
            \State $\tilde{\w}_a \gets \left(\XT\Y\right)_a \tilde{\q}_a$
            
        \Else \Comment{$2 \leq M \land K \leq M$}
            \State $\lambda_a \gets \text{largest eigenvalue of } \left(\XT\Y\YT\X\right)_a$
            \State $\tilde{\w}_a \gets \text{eigenvector corresponding to } \lambda_a$
        \EndIf
    \EndIf
    \State $\w_a \gets \frac{\tilde{\w}_a}{\norm{\tilde{\w}_a}}$
    \end{algorithmic}
\end{algorithm}

Then, step 3 consists of computing $\vr_a$, which can be done as shown in \mysecref{sec:improvement_r}, and, in step 4, \citet{dayal1997improved} compute $\vp_a$ and $\q_a$ as shown in \myalgref{alg:step_4_original}.

\begin{algorithm}
    \caption{Step 4 of IKPLS}\label{alg:step_4_original}
    \begin{algorithmic}[1]
    \If{algorithm $=1$} \Comment{IKPLS algorithm 1}
        \State $\t_a \gets \X \vr_a$
        \State $\vp_a \gets \frac{\XT\t_a}{\normsq{\t_a}}$
    \Else \Comment{IKPLS algorithm 2}
        \State $\normsq{\t_a} \gets \rT_a\XT\X\vr_a$ \Comment{$\rT_a\XT\X\vr_a=\tT_a\t_a$}
        \State $\vp_a \gets \frac{\XT\X\vr_a}{\normsq{\t_a}}$
    \EndIf
    \State $\q_a \gets \frac{\left(\rT_a\left(\XT\Y\right)_a\right)^\bTR}{\normsq{\t_a}}$
    \end{algorithmic}
\end{algorithm}

Note that line 5 of \myalgref{alg:step_4_original} produces $\normsq{\t_a}$ without computing $\t_a$. Substitute the definition of $\t_a$ from line 2 into line 5 to see this. This allows reference to $\normsq{\t_a}$ regardless of whether IKPLS algorithm 1 or 2 is used.

Now, consider \myalgref{alg:step_4_improved}, which is an improvement of \myalgref{alg:step_4_original}. The proportionality between $\q_a$ and $\tilde{\q}_a$ exploited by line~12 of \myalgref{alg:step_4_improved} is posed as a question--- ``is \texttt{q} proportional to \texttt{q.a}?'', with \texttt{q} denoting $\tilde{\q}_a$ and \texttt{q.a} denoting $\q_a$--- in a source comment in the function \texttt{kernelpls.fit} of the R package \texttt{pls} \citep[version~2.9-0]{liland2026pls, mevik2007the}. However, it nevertheless computes $\q_a$ as in line~8 of \myalgref{alg:step_4_original}. The comment has been present since version~2.0-0 (2006). \mytheoremref{theorem:q} answers the question affirmatively and supplies the constant of proportionality. In the same function, the $M = 1$ branch also computes $\q_a$ as in line~8, although the divisor it uses to normalize $\tilde{\w}_a$ into $\w_a$ is exactly the numerator in line~9 of \myalgref{alg:step_4_improved}.

\begin{algorithm}
    \caption{Improved step 4 of IKPLS}\label{alg:step_4_improved}
    \begin{algorithmic}[1]
    \If{algorithm $=1$} \Comment{IKPLS algorithm 1}
        \State $\t_a \gets \X \vr_a$
        \State $\vp_a \gets \frac{\XT\t_a}{\normsq{\t_a}}$
    \Else \Comment{IKPLS algorithm 2}
        \State $\normsq{\t_a} \gets \rT_a\XT\X\vr_a$ \Comment{$\rT_a\XT\X\vr_a=\tT_a\t_a$}
        \State $\vp_a \gets \frac{\XT\X\vr_a}{\normsq{\t_a}}$
    \EndIf
    \If{$M = 1$} \Comment{PLS1}
        \State $\q_a \gets \frac{\norm{\tilde{\w}_a}}{\normsq{\t_a}}$ \Comment{$\norm{\tilde{\w}_a}$ from \myalgref{alg:step_2}.}
    \Else \Comment{PLS2}
        \If{$M < K$} \Comment{$2 \leq M < K$}
            \State  $\q_a \gets \frac{\sqrt{\lambda_a}\tilde{\q}_a}{\norm{\tilde{\q}_a}\normsq{\t_a}}$ \Comment{$\lambda_a$ and $\tilde{\q}_a$ from \myalgref{alg:step_2}.}
        \Else \Comment{$2 \leq M \land K \leq M$}
            \State $\q_a \gets \frac{\left(\rT_a\left(\XT\Y\right)_a\right)^\bTR}{\normsq{\t_a}}$
        \EndIf
    \EndIf
    \end{algorithmic}
\end{algorithm}

\subsection{Correctness}
In this subsection, I will prove the correctness of \myalgref{alg:step_4_improved} by showing that it always computes the same $\q_a$ as \myalgref{alg:step_4_original}.

Consider step 5 of the IKPLS algorithms, as it is necessary to establish equivalence between \myalgref{alg:step_4_original} and \myalgref{alg:step_4_improved}. Step 5 deflates $\left(\XT\Y\right)_a$ and is given by \myeqnref{eq:step_5}.

\begin{equation}\label{eq:step_5}
    \left(\XT\Y\right)_{a+1} = \left(\XT\Y\right)_a - \vp_a\qT_a\normsq{\t_a}
\end{equation}

Furthermore, it is assumed that the extraction of PLS components halts if $\left(\XT\Y\right)_a$ becomes a zero matrix or if $\t_a=\zeros_N$: in either case, all covariance between $\X$ and $\Y$ has been extracted, and extraction stops with $a-1$ being the final component. The following lemmas can now be established.

\begin{lemma}\label{lemma:PTR}
    $\PT_a \R_a = \I$ where $\I$ is the identity matrix with $a$ rows and columns.

    \begin{proof}
        Consider an index $(i,j)$ into $\PT_a \R_a$. It is given by

        \begin{equation*}
            \begin{split}
                \left(\PT_a \R_a\right)_{i,j} &= \pT_i\vr_j\\
                &= \frac{\tT_i\X\vr_j}{\normsq{\t_i}}\\
                &= \frac{\tT_i\t_j}{\normsq{\t_i}}\\
                &= \frac{\tT_i\t_j}{\tT_i\t_i}\\
                &= \delta_{i,j}.
            \end{split}
        \end{equation*}
        The last equality follows from the column vectors of $\T$ being mutually orthogonal (\citealt{hoskuldsson1988pls}), and $\normsq{\t_i}>0$ holds for any extracted non-zero score, $\t_i$.
    \end{proof}
\end{lemma}

\begin{lemma}\label{lemma:r_zero}
    Consider an execution of IKPLS in which the $\q_i$ consumed by each performed deflation at index $i$ satisfies $\q_i\normsq{\t_i}=\left(\rT_i\left(\XT\Y\right)_i\right)^\bTR$. Then $\left(\XT\Y\right)^{\bTR}_a \vr_i = \zeros_M$ for every component $a$ of the execution and all $1 \leq i \leq a-1$.

    \begin{proof}
        Equivalently, by transposing, the conclusion states that $\rT_i\left(\XT\Y\right)_a = \zerosT_M$ for all $1 \leq i \leq a-1$. The supposition is in force throughout; the proof is by induction on $a$ over the conclusion alone.

        \emph{Base step ($a = 1$).} The conclusion quantifies over the empty range $1 \leq i \leq 0$ and is therefore vacuously true.

        \emph{Inductive step ($a \geq 2$).} Assume, as the inductive hypothesis, the conclusion for $a-1$:

        \begin{equation}\label{eq:ri_inductive_hypothesis}
            \rT_i\left(\XT\Y\right)_{a-1} = \zerosT_M \text{ for all } 1 \leq i \leq a-2.
        \end{equation}

        To show the conclusion for $a$, fix $i$ with $1 \leq i \leq a-1$. Left-multiplying \myeqnref{eq:step_5} by $\rT_i$ yields

        \begin{equation}\label{eq:ri_recurrence}
            \begin{split}
                \rT_i\left(\XT\Y\right)_{a} &= \rT_i\left(\XT\Y\right)_{a-1} - \rT_i\vp_{a-1}\qT_{a-1}\normsq{\t_{a-1}}\\
                &= \rT_i\left(\XT\Y\right)_{a-1} - \delta_{a-1,i}\qT_{a-1}\normsq{\t_{a-1}},
            \end{split}
        \end{equation}

        where the second equality follows from \mylemmaref{lemma:PTR}. Two cases remain.

        If $i \leq a-2$, then $\delta_{a-1,i}=0$, and the remaining term is $\zerosT_M$ by the inductive hypothesis, \myeqnref{eq:ri_inductive_hypothesis}, so $\rT_i\left(\XT\Y\right)_a = \zerosT_M$.

        If $i = a-1$, then $\delta_{a-1,i}=1$, and the supposition of \mylemmaref{lemma:r_zero}, at $i=a-1$ and transposed, gives $\qT_{a-1}\normsq{\t_{a-1}} = \rT_{a-1}\left(\XT\Y\right)_{a-1}$. Hence

        \begin{equation*}
            \begin{split}
                \rT_{a-1}\left(\XT\Y\right)_a &= \rT_{a-1}\left(\XT\Y\right)_{a-1} - \rT_{a-1}\left(\XT\Y\right)_{a-1}\\
                &= \zerosT_M.
            \end{split}
        \end{equation*}

        In both cases, $\rT_i\left(\XT\Y\right)_a = \zerosT_M$, completing the induction.
    \end{proof}
\end{lemma}

\begin{theorem}[Correctness, \myalgref{alg:step_4_improved}]\label{theorem:q}
    The value of $\q_a$ computed by \myalgref{alg:step_4_improved} equals the value of $\q_a$ computed by \myalgref{alg:step_4_original}.

    \begin{proof}
    \myalgref{alg:step_4_original} and \myalgref{alg:step_4_improved} differ only in the computation of $\q_a$. Line~8 of \myalgref{alg:step_4_original} always computes $\q_a$ as

        \begin{equation}\label{eq:q_alg2}
            \q_a = \frac{\left(\rT_a\left(\XT\Y\right)_a\right)^{\bTR}}{\normsq{\t_a}} = \frac{\left(\XT\Y\right)_a^{\bTR}\vr_a}{\normsq{\t_a}}.
        \end{equation}

        I prove, by strong induction on $a$, that the $\q_a$ computed by \myalgref{alg:step_4_improved} also satisfies \myeqnref{eq:q_alg2} for every $a \geq 1$; the induction enters only through \mylemmaref{lemma:r_zero}, in relating $\vr_a$ to $\w_a$. Consider \myeqnref{eq:r_improved} for computing $\vr_a$, and left-multiply it by $\left(\XT\Y\right)_a^{\bTR}$ to obtain the numerator in \myeqnref{eq:q_alg2}.

        For $a=1$, \myeqnref{eq:r_improved} contains no subtraction term, so, with nothing assumed,

        \begin{equation*}
            \left(\XT\Y\right)_1^{\bTR}\vr_1 = \left(\XT\Y\right)_1^{\bTR}\w_1.
        \end{equation*}

        For $a \geq 2$, assume, as the inductive hypothesis, that the $\q_i$ computed by \myalgref{alg:step_4_improved} satisfies \myeqnref{eq:q_alg2} at every index $i$ with $1 \leq i \leq a-1$. The deflations performed before component $a$, at indices $1, \ldots, a-1$, consumed precisely these $\q_i$, and multiplying \myeqnref{eq:q_alg2} by $\normsq{\t_i} > 0$ shows that each satisfies $\q_i\normsq{\t_i} = \left(\XT\Y\right)_i^{\bTR}\vr_i$. The execution up to the formation of $\left(\XT\Y\right)_a$ therefore satisfies the supposition of \mylemmaref{lemma:r_zero}, whose conclusion yields $\left(\XT\Y\right)_a^{\bTR}\vr_i = \zeros_M$ for all $1 \leq i \leq a-1$. Hence,

        \begin{equation*}
            \begin{aligned}
                &\left(\XT\Y\right)_a^{\bTR}\vr_a\\
                = &\left(\XT\Y\right)_a^{\bTR}\w_a - \left(\XT\Y\right)_a^{\bTR} \R_{a-1} \left(\PT_{a-1} \w_a\right)\\
                = & \left(\XT\Y\right)_a^{\bTR}\w_a,
            \end{aligned}
        \end{equation*}

        where the second equality holds by \mylemmaref{lemma:r_zero}.

        Thus, for all $a \geq 1$,

        \begin{equation*}
            \left(\XT\Y\right)_a^{\bTR}\vr_a = \left(\XT\Y\right)_a^{\bTR}\w_a.
        \end{equation*}

        Therefore, it suffices to show that each branch of \myalgref{alg:step_4_improved} reduces to

        \begin{equation}\label{eq:q_alg2_reduced}
            \q_a = \frac{\left(\XT\Y\right)_a^{\bTR}\w_a}{\normsq{\t_a}} = \frac{\left(\XT\Y\right)_a^{\bTR}\tilde{\w}_a}{\norm{\tilde{\w}_a}\normsq{\t_a}},
        \end{equation}

        where the second equality comes from line 13 of \myalgref{alg:step_2}. Under the halting assumptions stated below \myeqnref{eq:step_5}, every extracted component has $\left(\XT\Y\right)_a$ different from a zero matrix and $\t_a \neq \zeros_N$; the former ensures $\norm{\tilde{\w}_a}>0$ and the latter ensures $\normsq{\t_a}>0$, making \myeqnref{eq:q_alg2_reduced} well-defined. We now consider the three cases: $M=1$, $2 \leq M < K$, and $2 \leq M \land K \leq M$.

        \textbf{a)} \textit{First, consider the case where $M=1$ (PLS1).}
        
        Since $M=1$, line 2 of \myalgref{alg:step_2} computes $\tilde{\w}_a = \left(\XT\Y\right)_a$, so
        \begin{equation*}
            \left(\XT\Y\right)_a^{\bTR}\tilde{\w}_a = \tilde{\w}_a^{\bTR}\tilde{\w}_a = \normsq{\tilde{\w}_a},
        \end{equation*}
        and \myeqnref{eq:q_alg2_reduced} becomes
        \begin{equation*}
            \q_a = \frac{\normsq{\tilde{\w}_a}}{\norm{\tilde{\w}_a}\normsq{\t_a}} = \frac{\norm{\tilde{\w}_a}}{\normsq{\t_a}},
        \end{equation*}
        which is line 9 of \myalgref{alg:step_4_improved}.

        \textbf{b)} \textit{Now, consider the case where $2 \leq M < K$ (PLS2 with $\X$ larger than $\Y$).}
        
        Lines 5--7 of \myalgref{alg:step_2} set $\lambda_a$ to the largest eigenvalue of $\left(\YT\X\XT\Y\right)_a = \left(\XT\Y\right)_a^{\bTR}\left(\XT\Y\right)_a$, set $\tilde{\q}_a$ to a corresponding eigenvector, and set $\tilde{\w}_a = \left(\XT\Y\right)_a\tilde{\q}_a$. Then
        \begin{equation}\label{eq:XTY_w_1<M<K}
            \left(\XT\Y\right)_a^{\bTR}\tilde{\w}_a = \left(\XT\Y\right)_a^{\bTR}\left(\XT\Y\right)_a\tilde{\q}_a = \lambda_a\tilde{\q}_a.
        \end{equation}
        As $\left(\XT\Y\right)_a^{\bTR}\left(\XT\Y\right)_a$ is positive semi-definite, $\lambda_a \geq 0$, and, for non-zero $\left(\XT\Y\right)_a$, $\lambda_a > 0$. Furthermore,
        \begin{equation}\label{eq:w_1<M<K}
            \begin{aligned}
                \norm{\tilde{\w}_a} &= \sqrt{\tilde{\q}_a^{\bTR}\left(\XT\Y\right)_a^{\bTR}\left(\XT\Y\right)_a\tilde{\q}_a}\\
                &= \sqrt{\lambda_a\tilde{\q}_a^{\bTR}\tilde{\q}_a} = \sqrt{\lambda_a\norm{\tilde{\q}_a}^2}\\
                &= \sqrt{\lambda_a}\norm{\tilde{\q}_a}.
            \end{aligned}
        \end{equation}
        Substituting \myeqnref{eq:XTY_w_1<M<K} and \myeqnref{eq:w_1<M<K} into \myeqnref{eq:q_alg2_reduced} yields
        \begin{equation*}
            \q_a = \frac{\lambda_a\tilde{\q}_a}{\sqrt{\lambda_a}\norm{\tilde{\q}_a}\normsq{\t_a}} = \frac{\sqrt{\lambda_a}\tilde{\q}_a}{\norm{\tilde{\q}_a}\normsq{\t_a}},
        \end{equation*}
        which is line 12 of \myalgref{alg:step_4_improved}.

        \textbf{c)} \textit{Now, consider the final case where $2 \leq M \land K \leq M$ (PLS2 with $\X$ smaller than $\Y$).}

        Line 14 of \myalgref{alg:step_4_improved} computes 
        \begin{equation*}
            \q_a \gets \frac{\left(\rT_a\left(\XT\Y\right)_a\right)^\bTR}{\normsq{\t_a}}
        \end{equation*}
        and is identical to line 8 of \myalgref{alg:step_4_original}.

        The nested \texttt{else} clauses partition the possibilities into $M = 1$, $2 \leq M < K$, and $2 \leq M \land K \leq M$, which are exhaustive and non-overlapping, and $\q_a$ agrees with \myalgref{alg:step_4_original} in all cases due to \textbf{a)}, \textbf{b)}, and \textbf{c)}. This establishes \myeqnref{eq:q_alg2} at index $a$ and completes the induction. Since line~8 of \myalgref{alg:step_4_original} computes $\q_a$ precisely by \myeqnref{eq:q_alg2}, the $\q_a$ computed by \myalgref{alg:step_4_improved} equals that of \myalgref{alg:step_4_original} at every component $a$.
    \end{proof}
\end{theorem}

\begin{corollary}\label{corollary:drop_in}
    Executing IKPLS with \myalgref{alg:step_4_improved} in place of \myalgref{alg:step_4_original} produces identical $\W$, $\P$, $\Q$, $\R$, and $\T$.

    \begin{proof}
        \myalgref{alg:step_4_original} and \myalgref{alg:step_4_improved} compute all quantities other than $\q_a$ --- in particular $\tilde{\w}_a$, $\w_a$, $\vr_a$, $\vp_a$, and $\normsq{\t_a}$ --- identically, and $\q_a$ influences later components only through the deflation in \myeqnref{eq:step_5}. The run of IKPLS using \myalgref{alg:step_4_improved} therefore coincides with the run using \myalgref{alg:step_4_original} component by component: the runs coincide up to the formation of $\left(\XT\Y\right)_1 = \XT\Y$, and whenever they coincide up to the formation of $\left(\XT\Y\right)_a$, they compute identical $\tilde{\w}_a$, $\w_a$, $\vr_a$, $\vp_a$, and $\normsq{\t_a}$, obtain identical $\q_a$ by \mytheoremref{theorem:q}, and hence coincide up to the formation of $\left(\XT\Y\right)_{a+1}$ through \myeqnref{eq:step_5}. By induction on the components, the two runs coincide and produce identical $\W$, $\P$, $\Q$, $\R$, and $\T$.
    \end{proof}
\end{corollary}

\subsection{Runtime}
In this subsection, I analyze the runtime of computing $\q_a$ in Algorithms \ref{alg:step_4_original} and \ref{alg:step_4_improved}. I prove that the latter requires $\Theta\left(K\right)$ times fewer operations than the former, except when $2 \leq M \land K \leq M$, in which case the two coincide.

\begin{proposition}\label{prop:runtime_q_orig}
    The computation of $\q_a$ in line 8 of \myalgref{alg:step_4_original} requires $\Theta\left(KM\right)$ operations when $\left(\XT\Y\right)_a$ has been precomputed.
    \begin{proof}
        Computing the vector-matrix product $\rT_a\left(\XT\Y\right)_a$ and its subsequent transposition requires $\Theta\left(KM\right)$ operations. The subsequent element-wise division by $\normsq{\t_a}$ requires $\Theta\left(M\right)$ operations. Thus, the total number of required operations is $\Theta\left(KM\right) + \Theta\left(M\right) = \Theta\left(KM\right)$.
    \end{proof}
\end{proposition}

\begin{proposition}\label{prop:runtime_q_improved_M=1}
    The computation of $\q_a$ in line 9 of \myalgref{alg:step_4_improved} requires $\Theta(M)=\Theta\left(1\right)$ operations.
    \begin{proof}
        Scalar division between $\norm{\tilde{\w}_a}$ and $\normsq{\t_a}$ requires $\Theta(1)$ operations. Since $M=1$ due to the satisfied condition in line 8 of \myalgref{alg:step_4_improved}, the cost becomes $\Theta(M)=\Theta\left(1\right)$.
    \end{proof}
\end{proposition}

\begin{proposition}\label{prop:runtime_q_improved_1<M<K}
    The computation of $\q_a$ in line 12 of \myalgref{alg:step_4_improved} requires $\Theta(M)$ operations.
    \begin{proof}
        Computing $\sqrt{\lambda_a}$ requires $\Theta(1)$ operations, since $\lambda_a$ is already available from line 5 of \myalgref{alg:step_2}. The subsequent vector-scalar product $\tilde{\q}_a\sqrt{\lambda_a}$ requires an additional $\Theta\left(M\right)$ operations. In the denominator, computing $\norm{\tilde{\q}_a}$ requires $\Theta\left(M\right)$ operations, and the subsequent scalar product $\norm{\tilde{\q}_a} \normsq{\t_a}$ requires an additional $\Theta\left(1\right)$ operations. Finally, element-wise division of the numerator by the denominator requires $\Theta\left(M\right)$ operations. Thus, the total cost is $2\Theta(1) + 3\Theta(M) = \Theta(M)$ operations.
    \end{proof}
\end{proposition}

\begin{proposition}\label{prop:runtime_q_improved_2<=M_and_K<=M}
    The computation of $\q_a$ in line 14 of \myalgref{alg:step_4_improved} requires $\Theta\left(KM\right)$ operations.
    \begin{proof}
        This is identical to the proof of \mypropref{prop:runtime_q_orig}.
    \end{proof}
\end{proposition}

\begin{proposition}\label{prop:runtime_q_overall_improvement}
        \myalgref{alg:step_4_improved} requires $\Theta\left(K\right)$ times fewer operations than \myalgref{alg:step_4_original} to compute $\q_a$ when $M=1$ (PLS1) or $2 \leq  M < K$ (PLS2 with $\Y$ having fewer columns than $\X$), and \myalgref{alg:step_4_improved} requires the same number of operations as \myalgref{alg:step_4_original} when $2 \leq M \land K \leq M$ (PLS2 with $\Y$ having at least as many columns as $\X$).
    \begin{proof}
        By \mypropref{prop:runtime_q_orig}, \myalgref{alg:step_4_original} always requires $\Theta\left(KM\right)$ operations to compute $\q_a$.\\
    
        If $M=1$, then, by \mypropref{prop:runtime_q_improved_M=1}, \myalgref{alg:step_4_improved} requires $\Theta\left(M\right)$ operations to compute $\q_a$, which is $K$ times fewer than \myalgref{alg:step_4_original}.\\

        If $2 \leq M < K$, then, by \mypropref{prop:runtime_q_improved_1<M<K}, \myalgref{alg:step_4_improved} requires $\Theta\left(M\right)$ operations to compute $\q_a$, which is $K$ times fewer than \myalgref{alg:step_4_original}.\\

        Otherwise, $2 \leq M \land K \leq M$ and, then, by \mypropref{prop:runtime_q_improved_2<=M_and_K<=M}, \myalgref{alg:step_4_improved} requires $\Theta\left(KM\right)$ operations to compute $\q_a$, which is the same number as \myalgref{alg:step_4_original}.\\
    \end{proof}
\end{proposition}

\section{Additional improvements of IKPLS}\label{sec:improvements_additional}
In addition to the other improvements discussed in this article, the \texttt{ikpls} package (\citealt{engstrom2024fast}) also optimizes the computation of \myalgref{alg:step_2}. These optimizations are not discussed here, as they rely on reusing precomputed matrix products and optimizing the order of multiplication in matrix products involving more than two matrices. For example, $\XT\Y\YT\X$ can be obtained quickly by using the precomputed $\XT\Y$ and the equivalence $\XT\Y\YT\X = \XT\Y\left((\XT\Y)^\bTR\right)$, together with the fact that transposition is fast. These optimizations are not algorithmic improvements but efficient implementations. Therefore, they will not be discussed further, and the interested reader is referred to the source code of \texttt{ikpls} (\citealt{engstrom2024fast}) which contains similar optimizations for many other steps in the IKPLS algorithms.

\section{Benchmarks}\label{sec:benchmarks}
This section presents benchmarks demonstrating how the improvements established in Sections \ref{sec:improvement_r} and \ref{sec:improvement_q} reduce the practical runtime of both IKPLS algorithms relative to the original formulation by \citet{dayal1997improved}. All benchmarks are made using Python\footnote{\url{https://www.python.org/}} version 3.14 with the NumPy (\citealt[version 2.5.1]{harris2020array}) and JAX (\citealt[version 0.10.2, CUDA version 12.9]{jax2018github}) implementations of IKPLS in \texttt{ikpls} (\citealt[version 6.1.2]{engstrom2024fast}) with float64 precision. The NumPy implementations were executed on an AMD Ryzen 9 5950X processor utilizing all 16 cores (32 threads). The JAX implementations were executed on an Nvidia GeForce RTX 3090 Ti. For the JAX implementation, JIT-compiled variants were benchmarked, and compile-time overhead was not included. For both NumPy and JAX implementations, an initial few runs were executed and their results discarded to initialize the cache and ensure that the benchmarks executed first are not at a disadvantage due to a cold cache.

\subsection{Improvements in isolation}

\begin{figure*}[h]
    \begin{subfigure}[t]{0.49\textwidth}
        \centering
        \includegraphics[width=\textwidth]{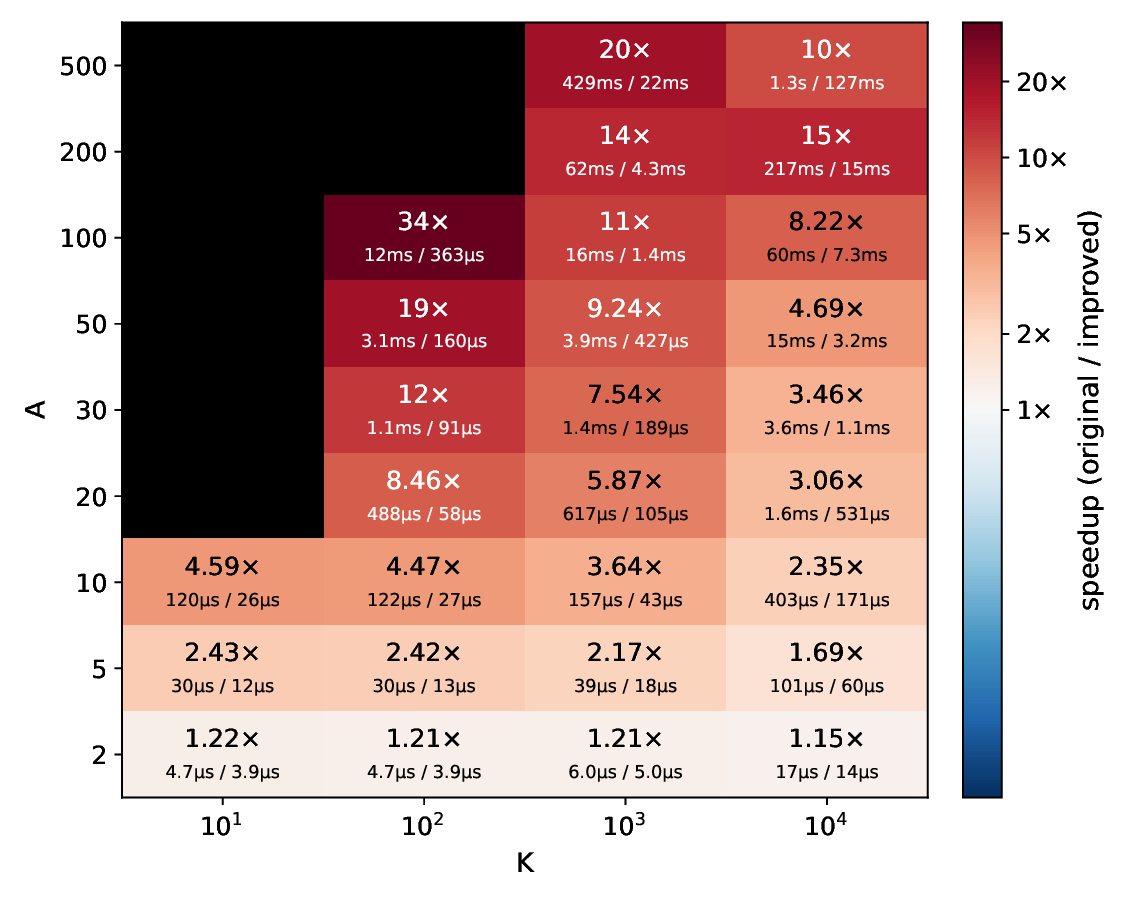}
        \caption{$\R$: speedup for $K \times A$ (NumPy, CPU).}
        \label{fig:R_heatmap_NumPy_CPU}
    \end{subfigure}
    \hfill
    \begin{subfigure}[t]{0.49\textwidth}
        \centering
        \includegraphics[width=\textwidth]{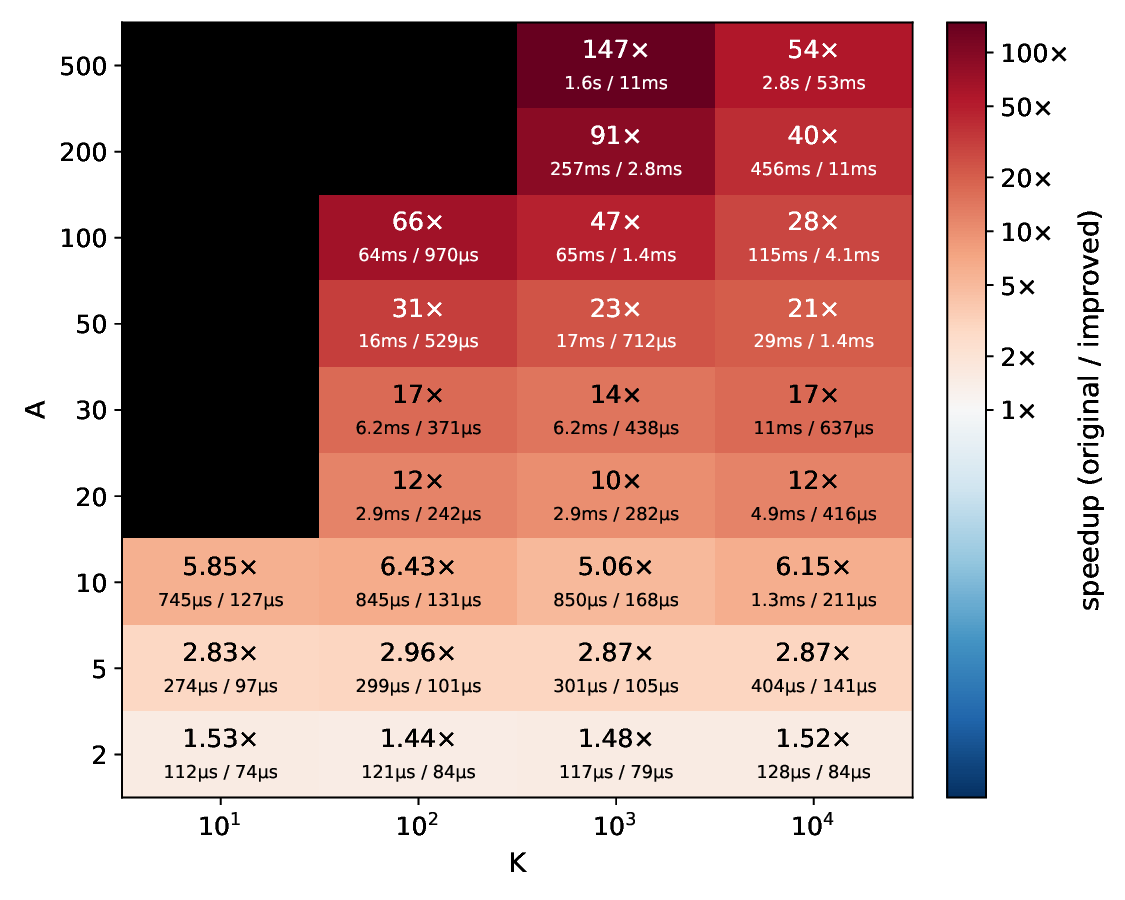}
        \caption{$\R$: speedup for $K \times A$ (JAX, GPU).}
        \label{fig:R_heatmap_JAX_GPU}
    \end{subfigure}
    \caption{Heatmap showing the speedup of the computation of $\R$ of shape $K \times A$ by using the evaluation strategy of \myeqnref{eq:r_improved} as opposed to the term-by-term accumulation strategy of \myeqnref{eq:r_original}. Since $A$ is upper-bounded by $\min (N,K)$, the cells with $K<A$ are not computed.}
    \label{fig:R_heatmaps}
\end{figure*}

The runtime of $\R$ is dependent on $K$ and $A$ as proven in Propositions \ref{prop:runtime_r_orig} and \ref{prop:runtime_r_improved}. Therefore, \myfigref{fig:R_heatmaps} shows practical runtimes for the computation of $\R$ with varying $K$ and $A$.

\begin{figure*}[h!t]
    \begin{subfigure}[t]{0.49\textwidth}
        \centering
        \includegraphics[width=\textwidth]{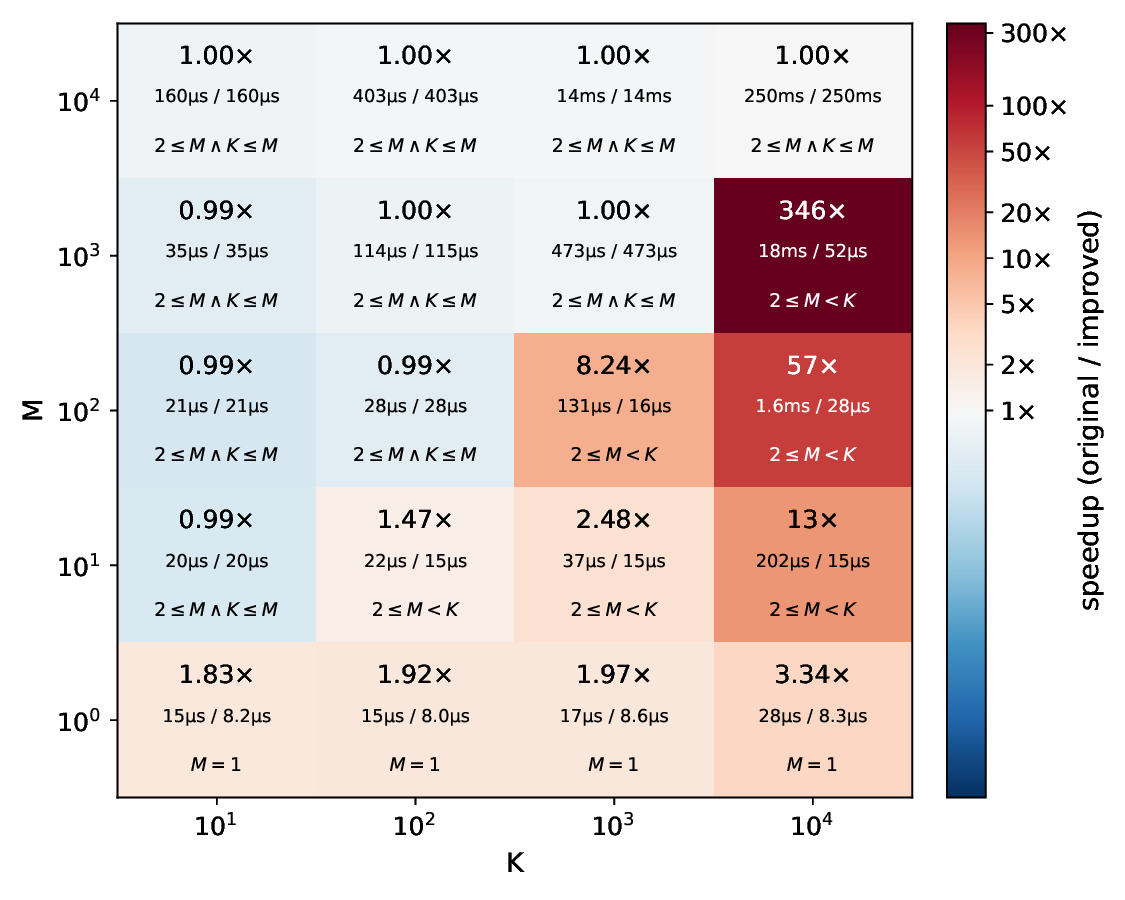}
        \caption{$\Q$: speedup for $K \times M$, $A=10$ (NumPy, CPU).}
        \label{fig:Q_heatmap_NumPy_CPU}
    \end{subfigure}
    \hfill
    \begin{subfigure}[t]{0.49\textwidth}
        \centering
        \includegraphics[width=\textwidth]{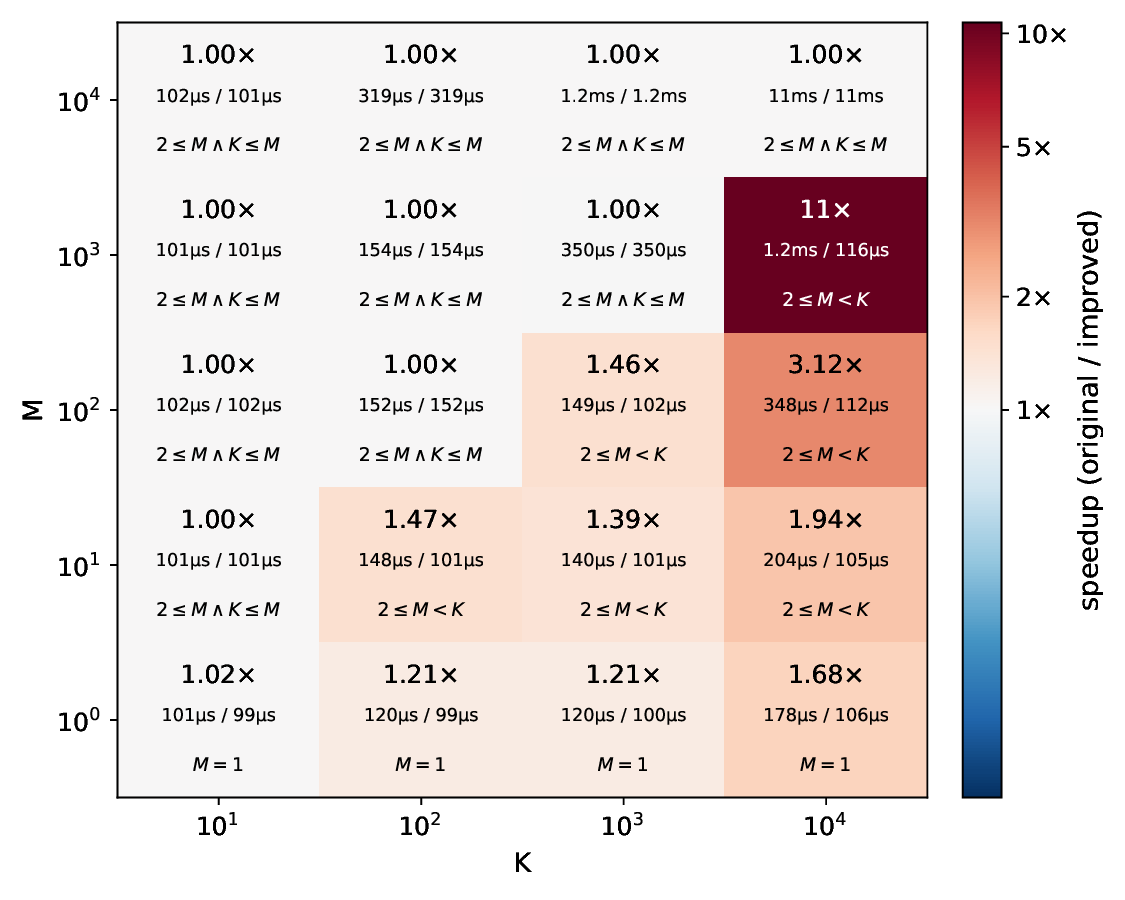}
        \caption{$\Q$: speedup for $K \times M$, $A=10$ (JAX, GPU).}
        \label{fig:Q_heatmap_JAX_GPU}
    \end{subfigure}
    \caption{Heatmap showing the speedup of the computation of $\Q$ of shape $M \times A$ with $A=10$ by using \myalgref{alg:step_4_improved} as opposed to \myalgref{alg:step_4_original}. Each $\q_a$ takes time independent of $a$, and so using $A=10$ simply takes $10$ times longer than computing any single $\q_a$.}
    \label{fig:Q_heatmaps}
\end{figure*}

Likewise, the runtime of $\q_a$ for any $a$ is dependent only on $K$ and $M$ as proven by Propositions \ref{prop:runtime_q_orig}, \ref{prop:runtime_q_improved_M=1}, \ref{prop:runtime_q_improved_1<M<K}, and \ref{prop:runtime_q_improved_2<=M_and_K<=M}. Therefore, \myfigref{fig:Q_heatmaps} shows practical runtimes for the computation of $\Q$ with varying $K$ and $M$.

Each benchmark in \myfigref{fig:R_heatmaps} and \myfigref{fig:Q_heatmaps} uses the same random seed and is the median of 30 to 1000 identical runs for each cell, to suppress variation from other factors. Faster cells receive more runs.

\myfigref{fig:R_heatmaps} reveals that the \textit{relative} speedup increases with $A$ and decreases with $K$ but that the new evaluation strategy of $\R$ is \textit{always} faster than the old one. My hypothesis is that as $A$ grows, the original evaluation strategy has to do an increasing amount of sequential work that the new evaluation strategy can parallelize. And as $K$ grows, the sequential steps themselves also grow, allowing better utilization of the parallel hardware within each step, thereby reducing the relative speedup. The relative speedups are generally most dramatic for the GPU, which is the more parallel of the two hardware types, corroborating the hypothesis.

\myfigref{fig:Q_heatmaps} corroborates \mypropref{prop:runtime_q_orig}, \mypropref{prop:runtime_q_improved_M=1}, \mypropref{prop:runtime_q_improved_1<M<K}, and \mypropref{prop:runtime_q_improved_2<=M_and_K<=M} nicely. For every value of $K$, the time required for computation of $\Q$ with \myalgref{alg:step_4_improved} is approximately constant when $M=1$ ($\approx 8\mu$s for the CPU and $\approx 100\mu$s for the GPU), and when $2 \leq M < K$, the time consumption generally increases with $M$ but is otherwise independent of $K$. When $2 \leq M \land K \leq M$, the time consumption increases with both $K$ and $M$ just like \myalgref{alg:step_4_original} does for all values of $K$ and $M$. For the GPU, when $K$ is small, even though $2 \leq M \land K \leq M$, the runtime does not seem to increase with increasing values of $M$. I hypothesize that this is an artifact of the GPU cores not being saturated at such small-sized operations.

Sometimes, when $2 \leq M \land K \leq M$, \myalgref{alg:step_4_improved} seems slightly slower than \myalgref{alg:step_4_original} when executed on a CPU. I attribute this to noise in the timing runs and, perhaps, to the fact that \myalgref{alg:step_4_improved} has to evaluate the conditionals on lines 8 and 11 before computing $\q_a$ in the same way as \myalgref{alg:step_4_original}, which does not have to evaluate the conditionals. In practice, this tiny overhead is completely negligible. In contrast, under JAX's just-in-time compilation, $M$ and $K$ are static properties of the input shapes, so the conditionals on lines~8 and~11 of \myalgref{alg:step_4_improved} are resolved when the function is traced, and only the selected branch is compiled: the hot runs execute no conditionals, making \myalgref{alg:step_4_original} and \myalgref{alg:step_4_improved} identical compiled programs when $2 \leq M \land K \leq M$. The dispatch that selects the compiled program from the input shapes at call time incurs the same cost for both algorithms, since compiled programs are specialized to the input shapes regardless of branching.

\subsection{Improvements relative to full fit}
\begin{figure*}[h!t]
    \begin{subfigure}[t]{0.49\textwidth}
        \centering
        \includegraphics[width=\textwidth]{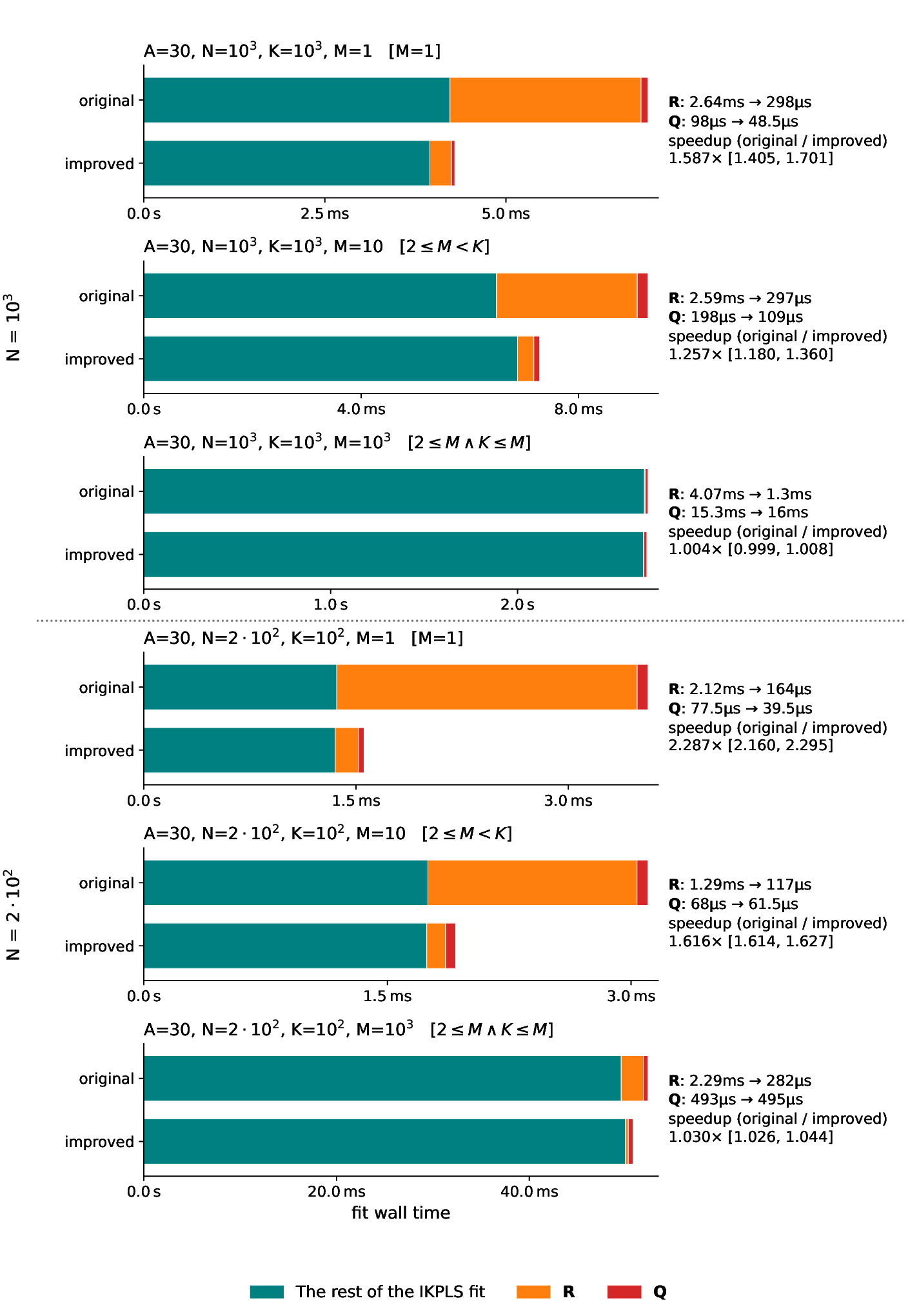}
        \caption{IKPLS Algorithm 1, NumPy, CPU.}
        \label{fig:alg1_fit_numpy}
    \end{subfigure}
    \hfill
    \begin{subfigure}[t]{0.49\textwidth}
        \centering
        \includegraphics[width=\textwidth]{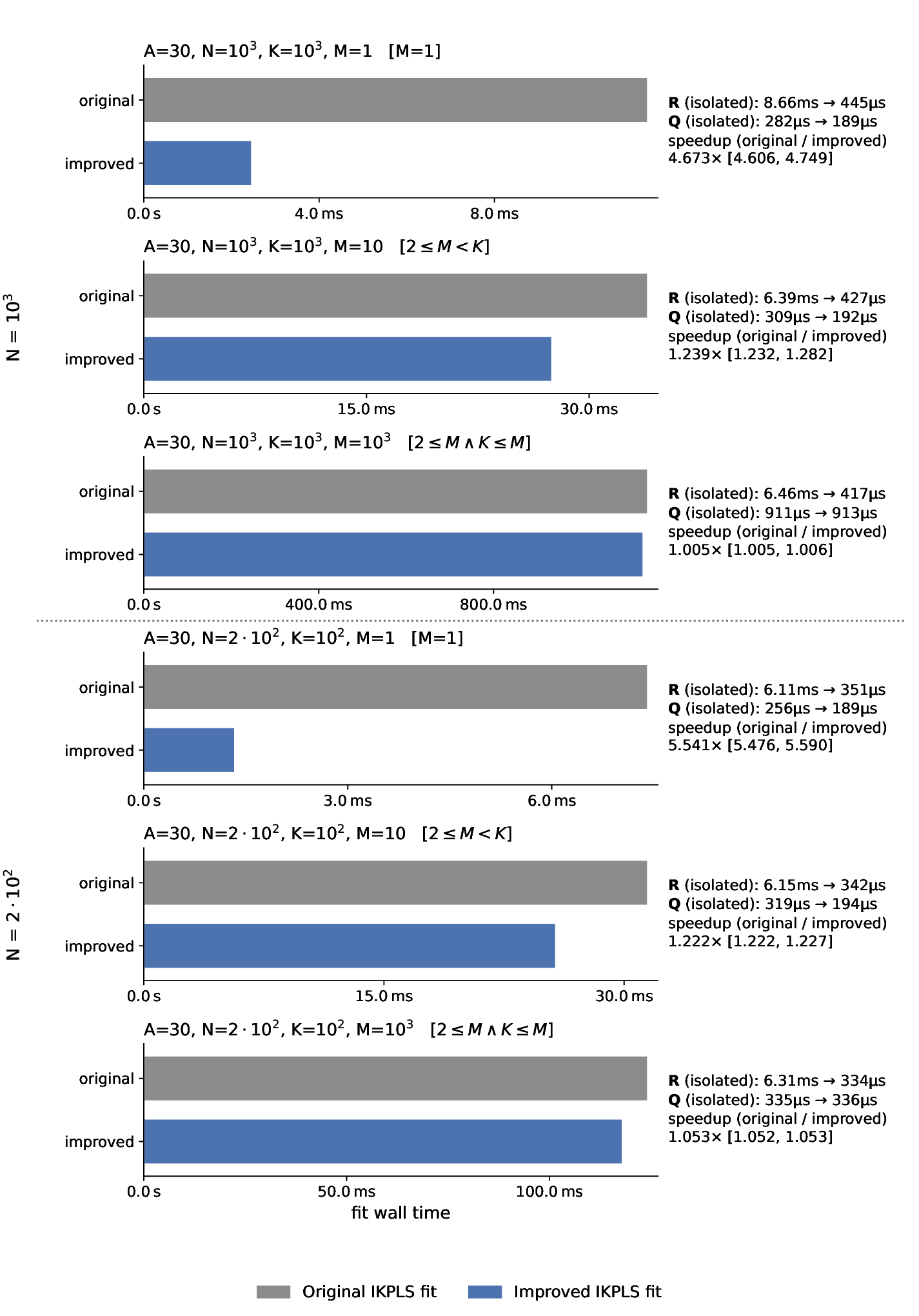}
        \caption{IKPLS Algorithm 1, JAX, GPU.}
        \label{fig:alg1_fit_jax}
    \end{subfigure}
    \\
    \begin{subfigure}[t]{0.49\textwidth}
        \centering
        \includegraphics[width=\textwidth]{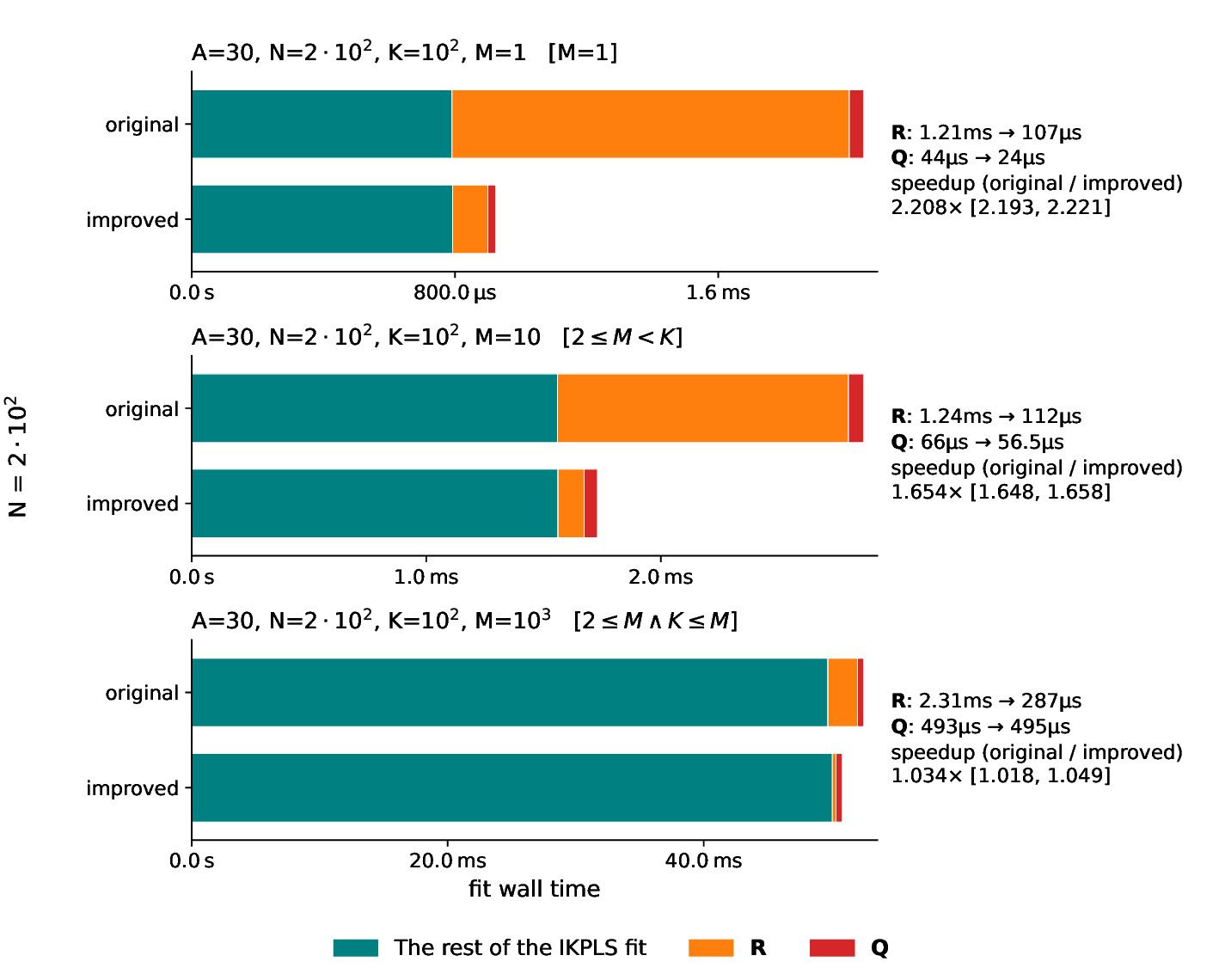}
        \caption{IKPLS Algorithm 2, NumPy, CPU.}
        \label{fig:alg2_fit_numpy}
    \end{subfigure}
    \hfill
    \begin{subfigure}[t]{0.49\textwidth}
        \centering
        \includegraphics[width=\textwidth]{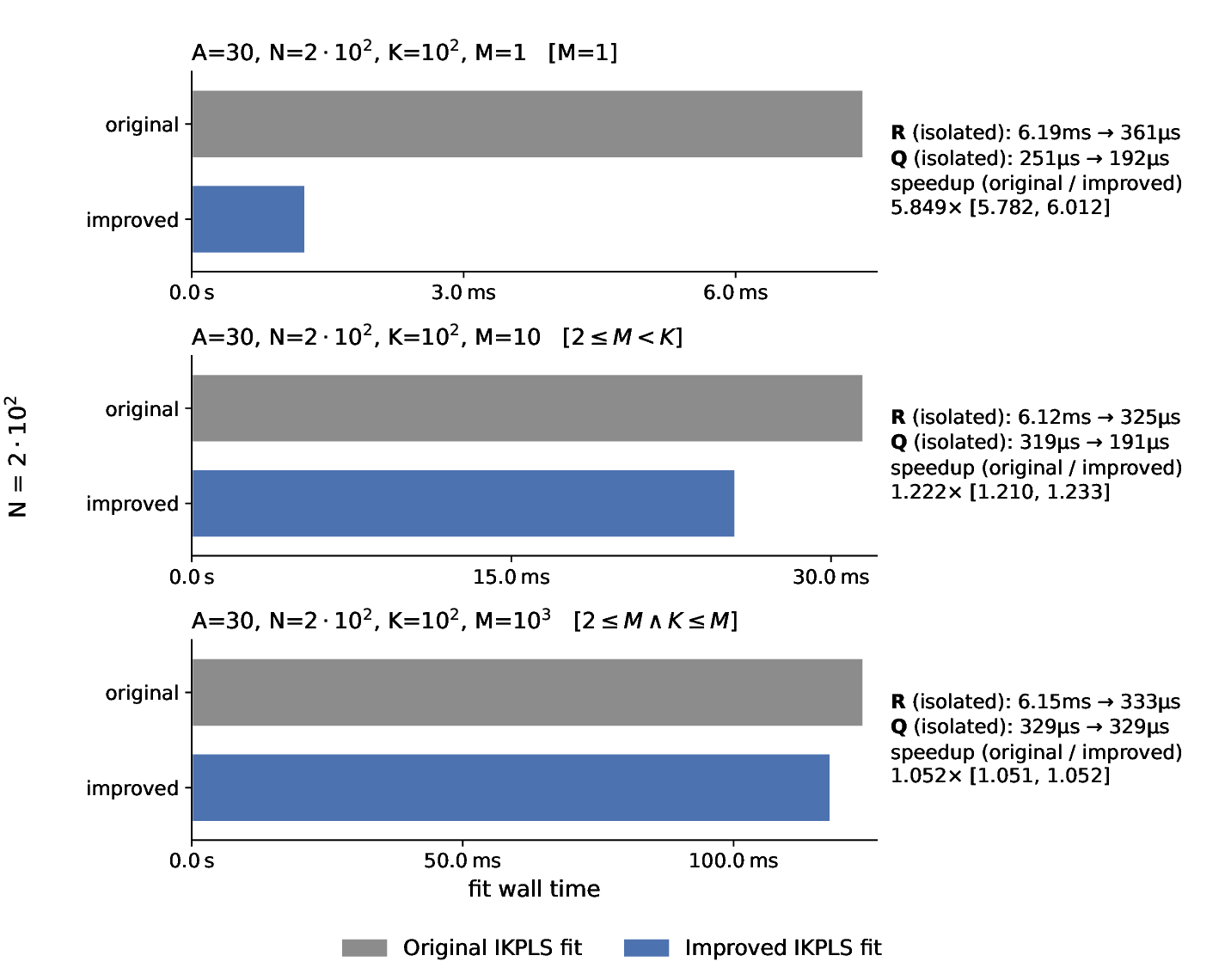}
        \caption{IKPLS Algorithm 2, JAX, GPU.}
        \label{fig:alg2_fit_jax}
    \end{subfigure}
    \caption{Time spent on full fits for the original and improved IKPLS algorithms. IKPLS algorithm 2 is only executed when $N > K$.\protect\footnotemark}
    \label{fig:full_fits}
\end{figure*}

\myfigref{fig:full_fits} shows a representative subset of the practical speedup achieved by the improved $\R$ and $\Q$ computations in the context of entire IKPLS fits. The speedups are estimated by the medians and the first and third quartiles across 10 runs using the original and improved implementations of both IKPLS algorithms.

For NumPy CPU implementations, the time spent on $\R$ and $\Q$ for the original and improved implementations is measured within the full fits. For the GPU runs, JAX compiles the entire fit routine into a fused XLA program, to the best of my knowledge, with no simple way to measure the wall-clock time spent on $\R$ and $\Q$. So, the runtime for $\R$ and $\Q$ is not measured directly during the fit but is instead estimated using compiled standalone programs that compute the same operations required by $\R$ and $\Q$, respectively. These isolated times are also each the median of 10 runs and include a roughly fixed CUDA kernel launch and JAX synchronization overhead that is independent of problem size and is not the in-fit contribution of $\R$ or $\Q$; only their reduction is meaningful.

The benchmarks are partitioned across $N=10^3$ as a surrogate for a reasonably sized dataset for PLS and $N=200$, which is relevant for either reasonably small datasets or local PLS modeling (\citealt{shenk1997investigation}) on a larger dataset. In all cases, the fits were done with $A=30$ components. The regime with $N < K$ was not explored, as practitioners concerned with speed will, in these cases, likely prefer the parsimonious kernel PLS (\citealt{liland2020much}) that computes the $N \times N$ $\X\XT$ matrix and operates with that, not unlike how IKPLS algorithm 2 operates on $\XT\X$.

Overall, the results are quite encouraging as the improved versions are faster than the originals in every case. The relative speedup is particularly large for the common PLS1 case ($M=1$). But there are also significant speedups achieved in the PLS2 case, most notably when $2 \leq M < K$. Generally, it seems that the improvement in $\R$ is most significant, although it was solely a practical consideration, whereas the improvement in $\Q$ decreased the asymptotic runtime. This is the case both for the actual in-fit time measurements on the CPU and the estimated values on the GPU.

It must be stated, however, that the relative speedups would be smaller for large $N$ as none of the optimizations depend on $N$ but other parts of the PLS fits do. Even when $M$ becomes large, especially when $M \geq K$, the relative speedups become vanishingly small as most of the time is then spent outside the computation of $\R$ and, if $M \geq K$, $\Q$ falls back to the original implementation.

\footnotetext{The observant reader may notice discrepancies between the relative speedups of $\R$ and $\Q$ in Figures \ref{fig:R_heatmaps} and \ref{fig:Q_heatmaps} and those in \myfigref{fig:full_fits}. This is likely because, within a full fit, the surrounding computation perturbs the cache residency and branch-predictor state under which $\R$ and $\Q$ execute.}

\section{Conclusion}\label{sec:conclusion}
This article identified improvements to the computations of $\R$ ($\X$ rotations) and $\Q$ ($\Y$ loadings) in both IKPLS algorithms. The $\R$ improvement replaces term-by-term accumulation with a parallel-friendly evaluation strategy requiring the same number of multiplications and is applicable whenever $A$ (the number of components) is at least $2$. Correctness is proved in Proposition~\ref{prop:r_correctness}, and the multiplication counts are established in Propositions~\ref{prop:runtime_r_orig}--\ref{prop:runtime_r_equal}.

The $\Q$ improvement (\myalgref{alg:step_4_improved}) derives $\Y$ loadings from quantities already computed in step 2 of IKPLS and is applicable when $M = 1$ (PLS1) or $2 \leq M < K$ (PLS2 with $\Y$ having fewer columns than $\X$). In both cases, the improved computation requires a factor of $\Theta\left(K\right)$ fewer operations than the original computation. Correctness is proved in \mytheoremref{theorem:q}, Corollary~\ref{corollary:drop_in} shows that the improvement is a drop-in replacement producing identical $\W$, $\P$, $\Q$, $\R$, and $\T$, and the runtime analysis is given in Propositions~\ref{prop:runtime_q_orig}--\ref{prop:runtime_q_overall_improvement}.

Finally, benchmarks showed that both improvements are realized in practice on both CPU and GPU implementations; that, depending on dataset size and shape, they speed up entire PLS fits by up to $\approx 2 \times$ on a CPU and $\approx 6 \times$ on a GPU; and that, for realistic dataset sizes and component counts, the $\R$ improvement contributes the larger share of the speedup, although only the $\Q$ improvement lowers the asymptotic operation count.

For IKPLS users and implementers, the results are twofold: \mysecref{sec:improvement_r} proves an evaluation strategy that is already used, without statement or proof, in the source code of the R package \texttt{pls} (\citealt{liland2026pls, mevik2007the}), and \mytheoremref{theorem:q} answers, affirmatively, a question that its source has posed about PLS2 since 2006 while also covering the PLS1 case. As the replacement is drop-in, the improved computation of $\Q$ is an immediate candidate for adoption there and in other existing IKPLS implementations.

Both improvements are implemented in the free, open-source Python package \texttt{ikpls} \citep[version 6.1.2]{engstrom2024fast}.

\clearpage
\bibliographystyle{bib} 
\bibliography{bib}

\end{document}